\documentclass[11pt,a4paper]{article}
\usepackage{authblk}
\usepackage[hyperref]{emnlp2020}
\usepackage{times}
\usepackage{latexsym}
\usepackage{graphicx}
\usepackage{xspace}
\usepackage{subcaption}
\usepackage{makecell}

\usepackage{microtype}

\aclfinalcopy 

\newcommand{\our}{\textsc{CORD-NER}\xspace}

\title{Comprehensive Named Entity Recognition on CORD-19 \\ with Distant or Weak Supervision}

\author{
	\textbf{Xuan Wang}$^{1}$,
	\textbf{Xiangchen Song}$^{1}$,
	\textbf{Bangzheng Li}$^{1}$,
	\textbf{Yingjun Guan}$^{2}$,
	\textbf{Jiawei Han}$^{1}$ \\
	$^{1}$Department of Computer Science, University of Illinois at Urbana-Champaign \\
	$^{2}$School of Information Sciences, University of Illinois at Urbana-Champaign \\
	$^{1,2}$\texttt{\{xwang174,xs22,yingjun2,bl17@,hanj\}@illinois.edu}
}

\date{}

\begin{document}
	\maketitle
	\begin{abstract}
		We created this \our dataset with comprehensive named entity recognition (NER) on the COVID-19 Open Research Dataset Challenge (CORD-19) corpus (2020-03-13). 
		This \our dataset covers 75 fine-grained entity types: In addition to the common biomedical entity types (e.g., genes, chemicals and diseases), it covers many new entity types related explicitly to the COVID-19 studies (e.g., coronaviruses, viral proteins, evolution, materials, substrates and immune responses), which may benefit research on COVID-19 related virus, spreading mechanisms, and potential vaccines. \our annotation is a combination of four sources with different NER methods. The quality of \our annotation surpasses SciSpacy (over 10\% higher on the F1 score based on a sample set of documents), a fully supervised BioNER tool. Moreover, \our supports incrementally adding new documents as well as adding new entity types when needed by adding dozens of seeds as the input examples. We will constantly update \our based on the incremental updates of the CORD-19 corpus and the improvement of our system. 
	\end{abstract}

	\section{Introduction}
	Coronavirus disease 2019 (COVID-19) is an infectious disease caused by severe acute respiratory syndrome coronavirus 2 (SARS-CoV-2). The disease was first identified in 2019 in Wuhan, Central China, and has since spread globally, resulting in the 2019–2020 coronavirus pandemic.
	On March 16th, 2020, researchers and leaders from the Allen Institute for AI, Chan Zuckerberg Initiative (CZI), Georgetown University's Center for Security and Emerging Technology (CSET), Microsoft, and the National Library of Medicine (NLM) at the National Institutes of Health released the COVID-19 Open Research Dataset (CORD-19)\footnote{\url{https://www.kaggle.com/allen-institute-for-ai/CORD-19-research-challenge}} of scholarly literature about COVID-19, SARS-CoV-2, and the coronavirus group.
	
	Named entity recognition (NER) is a fundamental step in text mining system development to facilitate COVID-19 studies. There is a critical need for NER methods that can quickly adapt to all the COVID-19 related new types without much human effort for training data annotation. We created this \textbf{\our dataset\footnote{\url{https://xuanwang91.github.io/2020-03-20-cord19-ner/}}} with comprehensive named entity annotation on the CORD-19 corpus (2020-03-13). This dataset covers 75 fine-grained named entity types. \our is automatically generated by combining the annotation results from four sources. 
	In the following sections, we introduce the details of \our dataset construction. We also show some NER annotation results in this dataset.

	\begin{table*}[t]
		\small 
		\centering
		\begin{tabular}{c|c|c|c|c|c|c|c|c|c|c|c|c}
			\hline
			& \multicolumn{3}{c|}{\textbf{Gene}} & \multicolumn{3}{c|}{\textbf{Chemical}} & \multicolumn{3}{c|}{\textbf{Disease}} & \multicolumn{3}{c}{\textbf{Total}} \\
			\cline{2-13}
			& Prec & Rec & F1 & Prec & Rec & F1 & Prec & Rec & F1 & Prec & Rec & F1 \\
			\hline
			\makecell[c]{SciSpacy\\(BIONLP13CG)} & \textbf{91.48} & \textbf{82.06} & \textbf{86.51} & 64.66 & 39.81 & 49.28 & 8.11 & 2.75 & 4.11 & 76.36 & 53.59 & 62.98 \\
			\hline
			\makecell[c]{SciSpacy\\(BC5CDR)} & - & - & - & \textbf{86.97} & 51.86 & 64.69 & \textbf{80.31} & 59.65 & 68.46 & \textbf{82.40} & 54.57 & 65.66 \\
			\hline
			Ours & 82.14 & 74.68 & 78.23 & 82.93 & \textbf{75.22} & \textbf{78.89} & 75.73 & \textbf{68.42} & \textbf{71.89} & 81.29 & \textbf{73.65} & \textbf{77.28} \\
			\hline
		\end{tabular}
		\caption {Performance comparison on three major biomedical entity types in COVID-19 corpus.}
		\label{tab:eval}
	\end{table*}
	
	\begin{figure*}[t]
		\centering
		\begin{subfigure}[b]{\textwidth}
			\centering
			\includegraphics[width=0.9\textwidth]{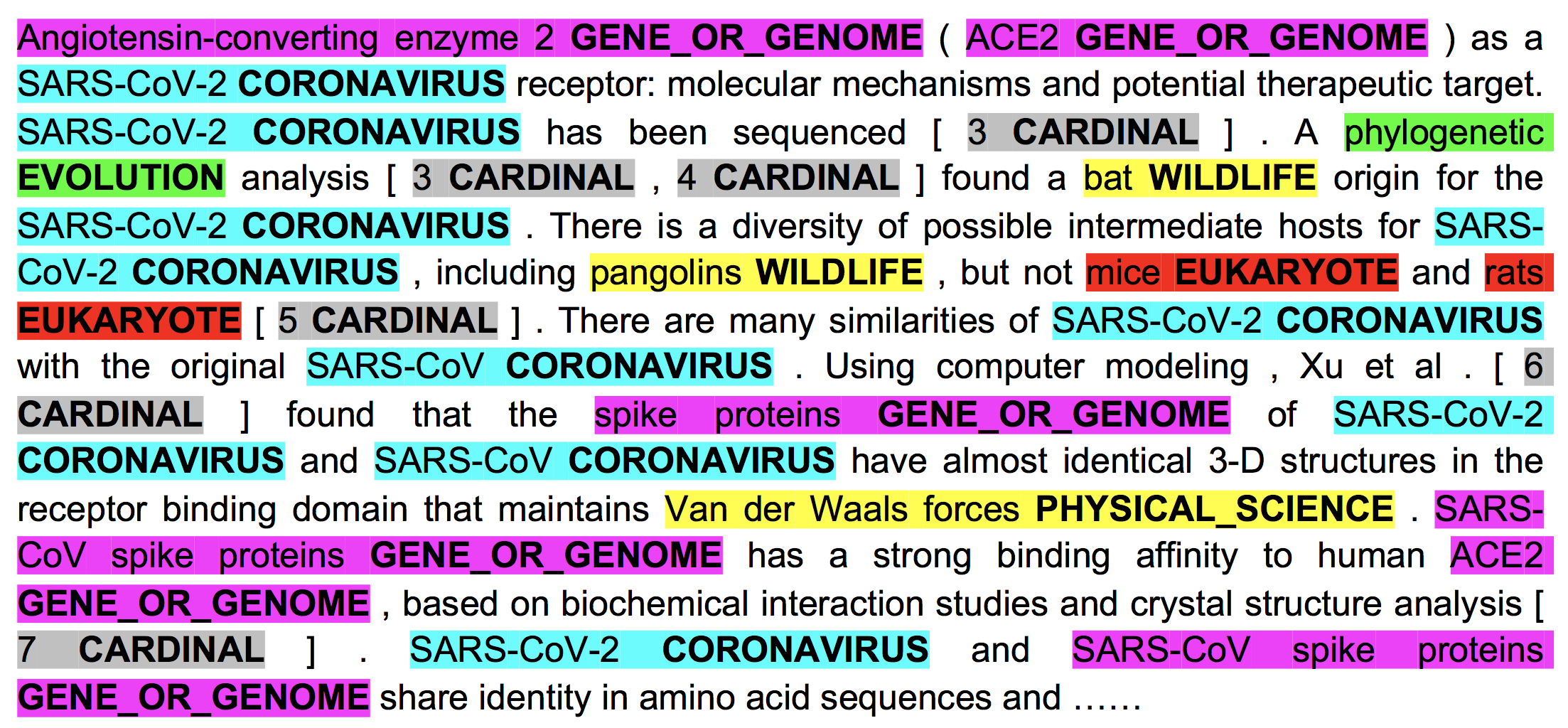}
			\caption{CORD-19 corpus}
			\label{subfig:annotation-cord}
		\end{subfigure}%
		
		\begin{subfigure}[b]{\textwidth}
			\centering
			\includegraphics[width=0.9\textwidth]{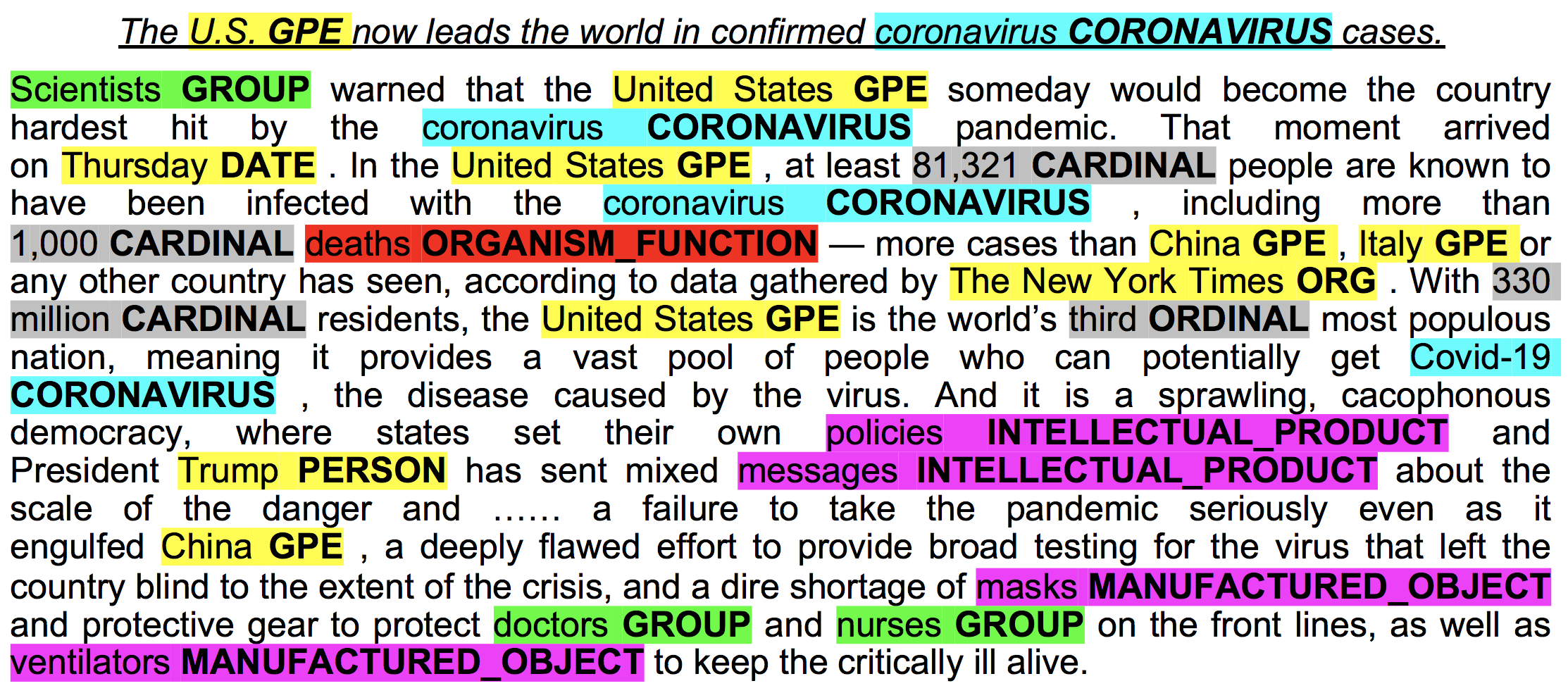}
			\caption{New York Times corpus}
			\label{subfig:annotation-nyt}
		\end{subfigure}%
		\caption{Examples of the annotation results with \our system.}
		\label{fig:annotation}
	\end{figure*}

	\begin{figure*}[t]
		\centering
		\begin{subfigure}[b]{\textwidth}
			\centering
			\includegraphics[width=0.9\textwidth]{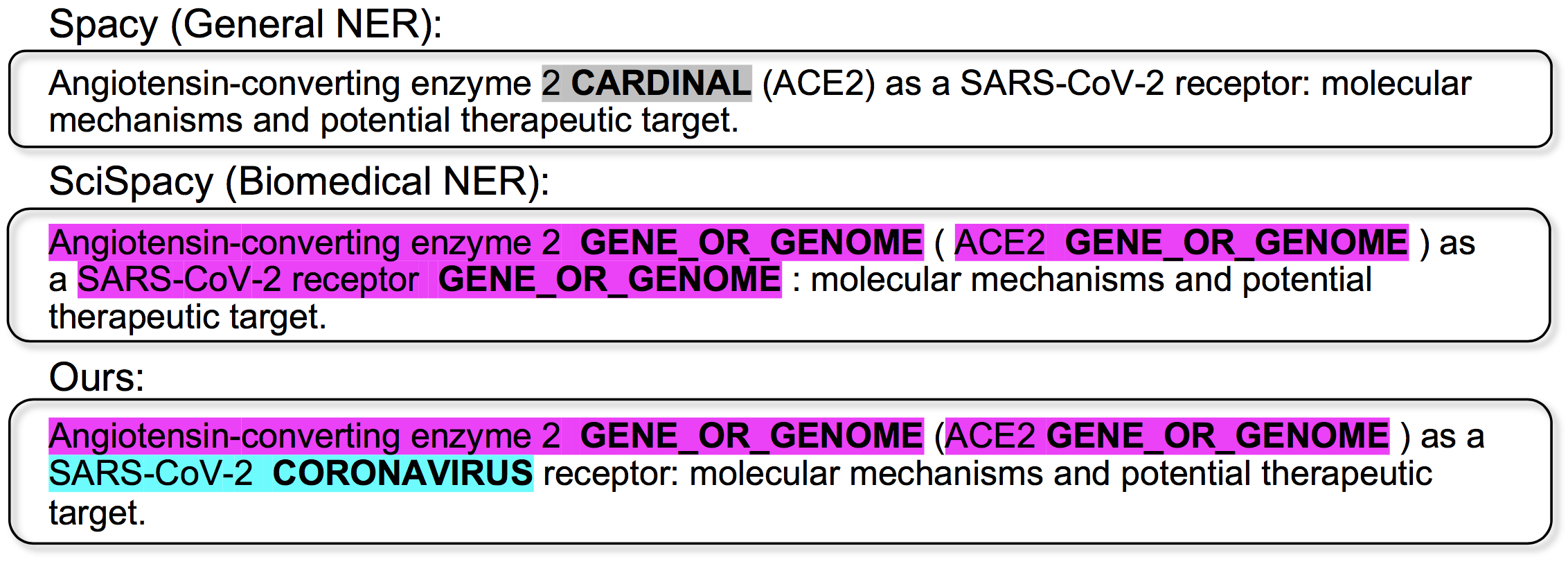}
			\caption{}
			\label{subfig:a}
		\end{subfigure}%
		
		\begin{subfigure}[b]{\textwidth}
			\centering
			\includegraphics[width=0.9\textwidth]{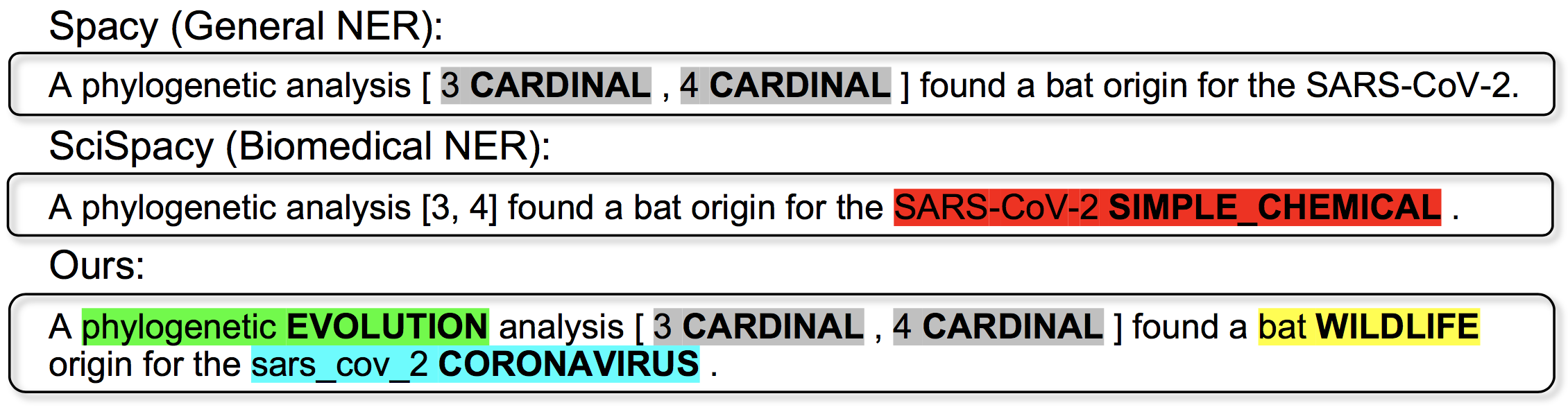}
			\caption{}
			\label{subfig:b}
		\end{subfigure}%
		
		\begin{subfigure}[b]{\textwidth}
			\centering
			\includegraphics[width=0.9\textwidth]{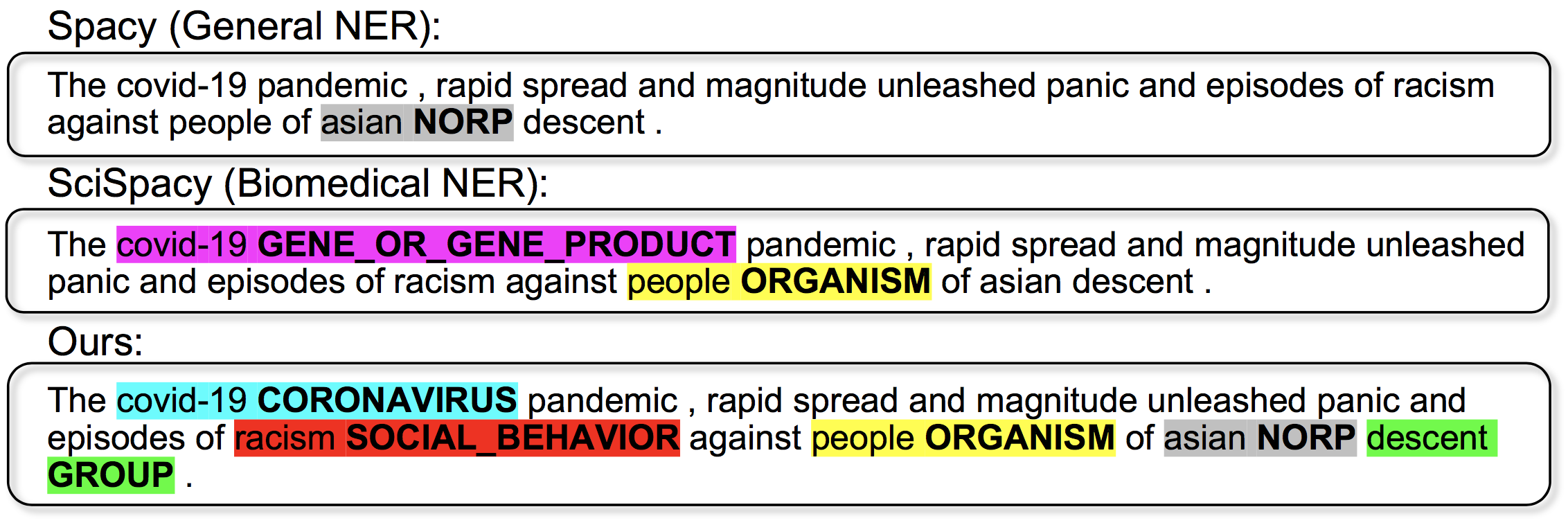}
			\caption{}
			\label{subfig:c}
		\end{subfigure}%
		\caption{Annotation result comparison with other NER methods.}
		\label{fig:annotation-comparison}
	\end{figure*}

	\section{\our Dataset}
	\subsection{Corpus}
	The input corpus is generated from the 29,500 documents in the CORD-19 corpus (2020-03-13). We first merge all the meta-data (all\_sources\_metadata\_2020-03-13.csv) with their corresponding full-text papers. Then we create a tokenized corpus (CORD-NER-corpus.json) for further NER annotations.
	
	The input corpus is a combination of the "title", "abstract" and "full-text" from the CORD-19 corpus. We first conduct automatic phrase mining and tokenization on the input corpus using AutoPhrase \cite{shang2018automated}. Then we do a second round of tokenization with Spacy\footnote{\url{https://spacy.io/api/annotation\#named-entities}} on the phrase-replaced corpus. We found that keeping the AutoPhrase results will significantly improve the distantly- and weakly-supervised NER performance.
	

	\subsection{NER Methods}
	\our annotation is a combination of four sources with different NER methods:
	\begin{enumerate}
		\item Pre-trained NER on 18 general entity types from Spacy using the model "en\_core\_web\_sm".
		\item Pre-trained NER on 18 biomedical entity types from SciSpacy\footnote{\url{https://allenai.github.io/scispacy/}} using the models ``en\_ner\_bionlp13cg\_md" and ``en\_ner\_bc5cdr\_md".
		\item Knowledgebase (KB)-guided NER on 127 biomedical entity types with our distantly-supervised NER methods \cite{wang2019distantly, shang2018learning}. We do not require any human-annotated training data for the NER model training. Instead, We rely on UMLS \footnote{\url{ttps://www.nlm.nih.gov/research/umls/META3_current_semantic_types.html}} as the input KB for distant supervision.
		\item Seed-guided NER on nine new entity types (specifically related to the COVID-19 studies) with our weakly-supervised NER method. We only require several (10-20) human-input seed entities for each new type. Then we expand the seed entity sets with CatE \cite{meng2020discriminative} and apply our distant NER method for the new entity type recognition.
	\end{enumerate}
	
	
	We reorganized all the entity types from the four sources into one entity type hierarchy (CORD-NER-types.xlsx). Specifically, we align all the types from SciSpacy to UMLS. We also merge some fine-grained UMLS entity types to their more coarse-grained types based on the corpus count. Our entity type hierarchy covers 75 fine-grained entity types: In addition to the common biomedical entity types (e.g., genes, chemicals and diseases), it covers many new entity types related explicitly to the COVID-19 studies (e.g., coronaviruses, viral proteins, evolution, materials, substrates and immune responses), which may benefit research on COVID-19 related virus, spreading mechanisms, and potential vaccines.
	
	Then we conduct named entity annotation on the 75 fine-grained entity types with the four sources of NER methods. After we get the NER annotation result with each method, we merge the results into one NER annotation file (CORD-NER.json). The conflicts are resolved by giving priority to different entity types annotated by different methods according to their annotation quality. Finally, we merge all the related information (meta-data, full-text corpus and NER results) into one file (CORD-NER-full.json) for users' convenience. The size of the dataset is about 1.2GB.

	\begin{table*}[t]
		\small 
		\centering
		\begin{tabular}{c|c|c|c}
			\hline
			CORONAVIRUS & EVOLUTION &  WILDLIFE & PHYSICAL SCIENCE \\
			\hline
			sars & mutation & bat & positively charged \\
			cov & phylogenetic & wild birds & negatively charged \\
			mers & evolution & wild animals & force field \\
			covid-19 & recombination & fruit bats & highly hydrophobic \\
			sars-cov-2 & substitutions & pteropus & van der waals interactions \\
			\hline\hline
			LIVESTOCK & MATERIAL & SUBSTRATE & IMMUNE\_ RESPONSE \\
			\hline
			pigs & air & blood & immunization \\
			poultry & plastic & urine & immunity \\
			calves & fluids & sputum & immune cells \\
			chicken & copper & saliva & innate immune \\
			pig & silica & fecal  & inflammatory response \\
			\hline\hline
			SIGN\_OR\_SYMPTOM & SOCIAL\_BEHAVIOR & INDIVIDUAL\_BEHAVIOR & \makecell[c]{THERAPEUTIC\_OR \\ \_PREVENTIVE \_PROCEDURE} \\
			\hline
			cough & collaboration & hand hygiene & detection \\
			respiratory symptoms & sharing & disclosures & vaccination \\
			diarrhoea & herd & absenteeism &  isolation \\
			vomiting & mediating & compliance & stimulation \\
			wheezing & adoption & empathy & inoculation \\
			\hline\hline
			DIAGNOSTIC\_PROCEDURE & RESEARCH\_ACTIVITY & EDUCATIONAL\_ACTIVITY & MACHINE\_ACTIVITY \\
			\hline
			imaging & rt-pcr & health education & machine learning \\
			immunohistochemistry & sequencing & workshops & data processing \\
			necropsy & screening & nursery &  automation \\
			scanning & diagnosis & medical education & deconvolution \\
			biopsy & prevention & residency & telecommunication \\
			\hline
		\end{tabular}
		\caption {Examples of the most frequent entities annotated in \our.}
		\label{tab:entities}
	\end{table*}

	\section{Results}
	\subsection{NER Annotation Results}
	In Table \ref{tab:eval}, we show the performance comparison between our annotation and the SciSpacy models. BIONLP13CG is the model in SciSpacy that covers the most entity types (18 entity types). BC5CDR is another model in SciSpacy that has the best performance on two entity types (chemicals and diseases). We manually annotated more than 1000 sentences for evaluation. Then we calculate the precision, recall and F1 scores on three major biomedical entity types: gene, chemical and disease. We can see that our annotation has worse performance on the gene type but much better performance on the chemical and disease types. In summary, the quality of our annotation surpasses SciSpacy by a large margin (over 10\% higher on the F1 score). Moreover, SciSpacy requires human effort for training data annotation and covers only 18 types. Our NER system supports incrementally adding new documents as well as adding new entity types when needed by adding dozens of seeds as the input examples.
	
	In Figure \ref{subfig:annotation-cord}, we show some examples of the annotation results in \our. We can see that our distantly- or weakly supervised methods achieve high quality recognizing the new entity types, requiring only several seed examples as the input. For instance, we recognized "SARS-CoV-2" as the "CORONAVIRUS" type, "bat" and "pangolins" as the "WILDLIFE" type and "Van der Waals forces" as the "PHYSICAL\_SCIENCE" type. This NER annotation can help downstream text mining tasks in discovering the origin and the physical nature of the virus. 
	Also, our NER methods are domain-independent that can be applied to the corpus in different domains. We show another example of NER annotation on the New York Times corpus with our system in Figure \ref{subfig:annotation-nyt}.
	
	In Figure \ref{fig:annotation-comparison}, we show the comparison of our annotation with existing fully-supervised NER/BioNER systems. In Figure \ref{subfig:a}, we can see that our method can identify "SARS-CoV-2" as a coronavirus. In Figure \ref{subfig:b}, we can see that our method can identify many more entities such as "phylogenetic" as an evolution term and "bat" as a wildlife term. In Figure \ref{subfig:c}, we can also see that our method can identify many more entities such as "racism" as social behavior. In summary, our distantly- and weakly-supervised NER methods are reliable for high-quality entity recognition without requiring human effort for training data annotation.

	\subsection{Top-Frequent Entity Summarization}
	In Table \ref{tab:entities}, we show some examples of the most frequent entities in our annotated corpus. Specifically, we show the entity types, including both our new types and some UMLS types that have not been manually annotated before.
	We find our annotated entities very informative for the COVID-19 studies. For example, the most frequent entities for the type "SIGN\_OR\_SYMPTOM behavior" includes "cough" and "respiratory symptoms" that are the most common symptoms for COVID-19. The most frequent entities for the type "INDIVIDUAL\_BEHAVIOR" include "hand hygiene", ``disclosures" and "absenteeism", which indicates that people focus more on hand cleaning for the COVID-19 issue. Also, the most frequent entities for the type "MACHINE\_ACTIVITY" include ``machine learning", ``data processing" and "automation", which indicates that people focus more on automated methods that can process massive data for the COVID-19 studies. This type also includes "telecommunication" as the top results, which is quite reasonable under the current COVID-19 situation. More examples can be found in our dataset.

	\section{Conclusion}
	\our will be constantly updated based on the incremental updates of the CORD-19 corpus and the improvement of our system. We will also build text mining systems based on the \our dataset with richer functionalities. We hope this dataset can help the text mining community build downstream applications for the COVID-19 related tasks. We also hope this dataset can bring insights for the COVID-19 studies on making scientific discoveries.

	\section*{Acknowledgment}
	Research was sponsored in part by US DARPA KAIROS Program No. FA8750-19-2-1004 and SocialSim Program No.  W911NF-17-C-0099, National Science Foundation IIS 16-18481, IIS 17-04532, and IIS-17-41317, and DTRA HDTRA11810026. Any opinions, findings, and conclusions or recommendations expressed herein are those of the authors and should not be interpreted as necessarily representing the views, either expressed or implied, of DARPA or the U.S. Government. The U.S. Government is authorized to reproduce and distribute reprints for government purposes notwithstanding any copyright annotation hereon. The views and conclusions contained in this paper are those of the authors and should not be interpreted as representing any funding agencies.

	\bibliography{anthology,emnlp2020}
	\bibliographystyle{acl_natbib}
	
\end{document}